\documentclass[conference]{IEEEtran}
\IEEEoverridecommandlockouts
\usepackage{cite}
\usepackage{amsmath,amssymb,amsfonts}
\usepackage{algorithmic}
\usepackage{graphicx}
\usepackage{textcomp}
\usepackage{xcolor}
\usepackage[utf8]{inputenc}
\usepackage{graphicx}
\usepackage{subcaption}
\usepackage{tabularx, makecell}  
\usepackage{siunitx}
\def\BibTeX{{\rm B\kern-.05em{\sc i\kern-.025em b}\kern-.08em
    T\kern-.1667em\lower.7ex\hbox{E}\kern-.125emX}}
\begin{document}

\title{Enhancing Fitness Movement Recognition with Attention Mechanism and Pre-Trained Feature Extractors}

%


\author{\IEEEauthorblockN{Shanjid Hasan Nishat\IEEEauthorrefmark{1}, Srabonti Deb\IEEEauthorrefmark{2}, Mohiuddin Ahmed\IEEEauthorrefmark{3}}
\textit{Department of Computer Science \& Engineering} \\
\textit{Rajshahi University of Engineering \& Technology, Rajshahi-6204, Bangladesh} \\
Emails: \IEEEauthorrefmark{1}shanjidhasan202@gmail.com, \IEEEauthorrefmark{2}srabontideb23@gmail.com, \IEEEauthorrefmark{3}mohiuddin.nirob.mn@gmail.com}
\maketitle
\begin{center}
\noindent
\fbox{
\begin{minipage}{0.95\linewidth}
\footnotesize
\textbf{Disclaimer:} This work has been submitted to the IEEE for possible publication. Copyright may be transferred without notice, after which this version may no longer be accessible.
\end{minipage}
}
\end{center}

\begin{abstract}

Fitness movement recognition, a focused subdomain of human activity recognition (HAR), plays a vital role in health monitoring, rehabilitation, and personalized fitness training by enabling automated exercise classification from video data. However, many existing deep learning approaches rely on computationally intensive 3D models, limiting their feasibility in real-time or resource-constrained settings. In this paper, we present a lightweight and effective framework that integrates pre-trained 2D Convolutional Neural Networks (CNNs) such as ResNet50, EfficientNet, and Vision Transformers (ViT) with a Long Short-Term Memory (LSTM) network enhanced by spatial attention. These models efficiently extract spatial features while the LSTM captures temporal dependencies, and the attention mechanism emphasizes informative segments. We evaluate the framework on a curated subset of the UCF101 dataset, achieving a peak accuracy of 93.34\% with the ResNet50-based configuration. Comparative results demonstrate the superiority of our approach over several state-of-the-art HAR systems. The proposed method offers a scalable and real-time-capable solution for fitness activity recognition with broader applications in vision-based health and activity monitoring.

\end{abstract}


\section{Introduction}
\IEEEpubidadjcol
Human activity recognition (HAR), a fundamental aspect of intelligent vision systems, enables computational interpretation of physical activities from video or sensor data. HAR continues to play a pivotal role in various domains including video surveillance~\cite{ref1}, healthcare~\cite{ref2}, human-computer interaction~\cite{ref3}, and autonomous driving~\cite{ref4}. A prominent subfield within this domain is fitness movement recognition, which focuses specifically on identifying exercise-related actions for applications such as personalized monitoring, real-time feedback, and rehabilitation support~\cite{ref5}. Accurate recognition of fitness activities contributes to improved health management and disease prevention~\cite{ref6}. Therapists emphasize the importance of activity-based interventions, and modern technological platforms such as wearable devices and tele-exercise systems facilitate these goals by offering real-time feedback and sustained user engagement~\cite{ref7}.

The progression of fitness technology reflects the growing shift toward data-driven solutions for promoting active lifestyles. From early pedometers to smartwatches and tele-exercise platforms, modern systems now offer features such as step counting, heart rate monitoring, and personalized exercise tracking~\cite{ref9}. Studies highlight their effectiveness in boosting short-term physical activity and reducing BMI, though sustained behavioral change remains underexplored~\cite{ref10}. Historically evolving from mechanical equipment to digital gym machines with metric displays~\cite{ref12}, today’s technologies leverage artificial intelligence and large-scale data to deliver adaptive and user-centric fitness insights.

While the importance of fitness movement recognition is widely acknowledged, existing computational models still face challenges. Traditional approaches relying on handcrafted features were limited in accuracy and scalability~\cite{ref15}. More recently, 3D Convolutional Neural Networks (CNNs)~\cite{cnn} have shown strong performance in activity recognition but suffer from high computational costs, which restrict their applicability in real-time and resource-constrained environments such as smartphones and wearable devices~\cite{ref16}. This limitation highlights the need for lightweight yet accurate frameworks capable of balancing performance with efficiency. To address these challenges, this research introduces a novel architecture that integrates pre-trained feature extractors with Long Short-Term Memory (LSTM)~\cite{lstm} networks enhanced by an attention mechanism. Specifically, pre-trained CNNs such as ResNet~\cite{resnet}, EfficientNet~\cite{efficientnet}, and Vision Transformers (ViT)~\cite{vit}, trained on large-scale datasets like ImageNet, are used to extract discriminative spatial features from fitness video frames. These features are then processed by an LSTM network to capture temporal dynamics, while the attention mechanism further improves performance by adaptively focusing on the most informative frames or features during recognition. The major contributions of this work include-
\begin{itemize}
\item Development of a lightweight hybrid framework integrating 2D CNN, LSTM, and attention mechanism for efficient fitness movement recognition, with significantly fewer parameters compared to 3D CNN-based models.

\item Comparative evaluation of multiple deep learning models to identify the most accurate and computationally efficient approach for fitness movement recognition.

\item Advancement of fitness monitoring technologies by providing a practical recognition system suitable for healthcare, personalized coaching, and rehabilitation.
\end{itemize}
The rest of the paper is structured as follows: Section II reviews key techniques and models in movement recognition. Section III outlines the proposed architecture of the model, dataset description, data preprocessing, and a detailed description of model components. Section IV discusses experimental setup, hyperparameters, and results. Finally, the last section provides the study's final analysis and prospective areas for future exploration.

\section{Related Works}
The field of HAR has progressed notably with the advent of deep learning, shifting from conventional handcrafted methods toward models that automatically extract spatial and temporal representations from video and sensor data. Early efforts such as those by Chen et al. \cite{chen2021} introduced supervised hashing frameworks for efficient video retrieval, emphasizing scalability through compact binary encodings. While such approaches were suitable for large-scale search tasks, they lacked the temporal modeling necessary for fine-grained motion recognition, especially in the fitness domain.

Vrskova et al. \cite{vrskova2022} proposed a lightweight 3D CNN that balances accuracy and computational cost. Despite its efficiency, the architecture was limited in capturing subtle motion patterns and overlapping actions. To overcome the challenges of generalization and temporal awareness, Zhou et al.~\cite{zhou2022} introduced a knowledge distillation method based on Vision Transformers, transferring spatial attention cues from a large supervisory model to a smaller one. Although this reduced model size, transformer-based frameworks typically require heavy pretraining and remain resource-intensive for real-time applications.

Huang et al.~\cite{huang2023} addressed class scalability through a zero-shot action recognition model using a dual-GAN to synthesize visual features of unseen classes from semantic embeddings. Yet, its performance declined on repetitive and temporally dependent motions typical in fitness contexts.

Focusing on explainability and privacy, Bsoul~\cite{bsoul2025} proposed a graph-based classifier leveraging skeletal keypoints. This offered interpretable pose-based predictions but was vulnerable to occlusions and noisy environments, limiting robustness. Dasan et al.~\cite{dasan2025} accelerated mobile HAR using MoViNet with reduced latency and model size, though the simplified design hindered the discrimination of visually similar exercises.

Maji et al.~\cite{maji2023} tackled low-latency activity recognition via a hybrid model combining convolutional layers with attention-augmented GRUs using wearable sensor input. Despite its temporal precision, reliance on sensor data constrains its adaptability for vision-only environments better suited for fitness tracking using camera feeds.

In light of these developments, our proposed model adopts a balanced hybrid architecture, integrating pre-trained 2D CNNs for spatial representation with LSTM layers for temporal dynamics, further enhanced by a spatial attention mechanism. Unlike zero-shot or sensor-based methods, our pipeline directly processes visual inputs, making it broadly applicable and independent of external sensing hardware. Moreover, it avoids the overhead of transformer-heavy models while retaining the benefit of attention-based focus on discriminative temporal segments. This design enables effective recognition of fine-grained fitness activities in real time, with improved generalization and scalability across environments.

\section{Materials and Methods} 
\begin{figure*}[htbp]
\centering
\includegraphics[width=\textwidth]{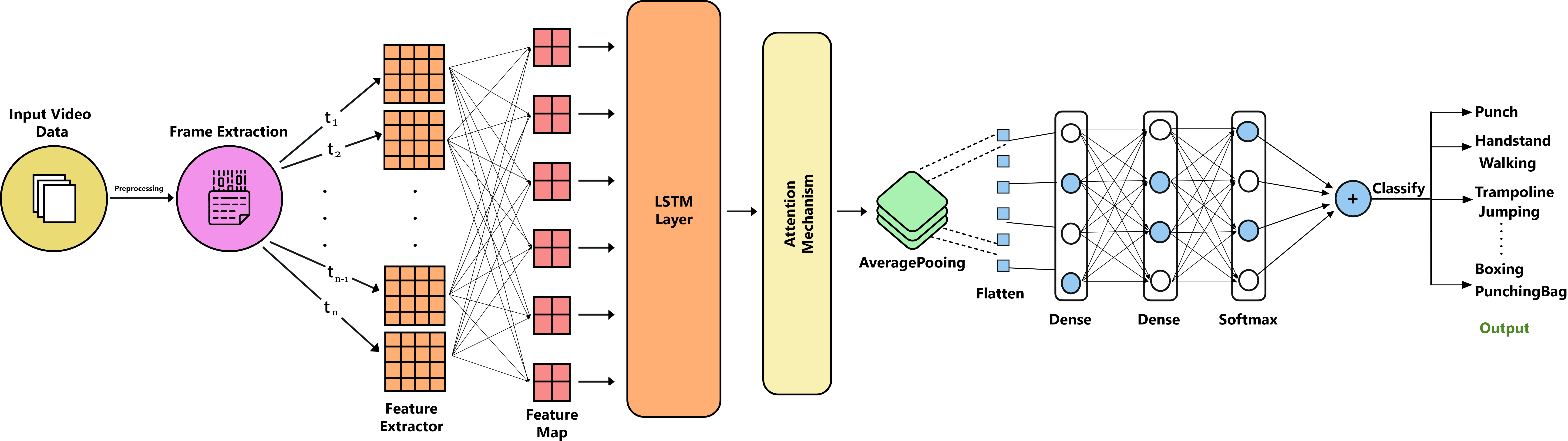}
\caption{Architecture of Proposed Methodology}
\label{fig1}
\end{figure*}

Our proposed architecture for fitness movement recognition is illustrated in Fig.\ref{fig1}. The pipeline is designed to effectively capture both spatial and temporal dynamics from video data, ensuring robust classification of exercise activities.

At the initial stage, a 2D feature extractor processed each frame of the input video sequence, capturing spatial patterns. These frame-wise features were passed through a TimeDistributed layer to maintain consistency across all frames. The extracted features were then flattened and fed into an LSTM layer, which models temporal dependencies and sequential dynamics. To enhance focus on the most informative regions, a spatial attention mechanism was applied to the LSTM output. The resulting attention weights were combined with the temporal features, emphasizing relevant spatial cues while suppressing irrelevant ones. The attended features were aggregated using global average pooling across the temporal dimension, producing a compact and discriminative representation. Finally, this representation was passed through a dense layer with softmax activation, enabling classification into the predefined fitness movement categories.

\subsection{Dataset Description}
In this study, we used the UCF101 dataset~\cite{soomro2012} to evaluate the performance of the proposed fitness movement recognition framework. UCF101 (University of Central Florida 101) is a widely adopted action recognition dataset consisting of 13,320 video clips categorized into 101 human activity classes. These classes span various domains, including sports, daily activities, and fitness-related exercises, making it suitable for human activity recognition research.

Each class in the dataset contains approximately 133 video clips on average, offering sufficient diversity and volume for training and evaluation. For this work, we selected a subset of fitness-related classes that align with the objectives of fitness movement recognition.
\subsection{Dataset Preprocessing}
To prepare the video data for model training, a standardized preprocessing pipeline was applied to ensure consistency across samples and compatibility with the deep learning architecture. The key steps include:

\begin{enumerate}
    \item \textit{Frame Extraction:} From each video, 20 frames were uniformly sampled at equal intervals to capture temporal dynamics. Given a video with $N$ frames, the selected frames $\{f_{t_1}, f_{t_2}, \dots, f_{t_{20}}\}$ were indexed using:
\begin{equation}
t_i = \left\lfloor \frac{i \times N}{21} \right\rfloor
\end{equation}

This sampling strategy maintains temporal consistency while reducing computational load.
    \item \textit{Frame Resizing:} All frames were initially resized to $64 \times 64$ for storage and processing efficiency. Before feature extraction using pre-trained models (trained on ImageNet with $224 \times 224$ inputs), frames were upscaled to $224 \times 224$ to ensure compatibility.
    \item \textit{Normalization:} Pixel values were normalized to the range [0, 1] by dividing each pixel intensity by 255:
    \begin{equation}
    \text{Normalized Value} = \frac{\text{Original Value}}{255}
    \end{equation}
\end{enumerate}

These preprocessing steps ensured a uniform and compact input representation, facilitating efficient training and stable convergence of the model.
\subsection{2D Feature Extractor}

To extract spatial features from individual video frames, a convolutional layer was employed as the foundational component. This layer applies a set of learnable filters (kernels) across the input image to detect local patterns such as edges, textures, and shapes. For a single-channel input image $X$ and a filter $W$, the convolution operation is expressed as:
\begin{equation}
F = X \ast W + b
\end{equation}
where $\ast$ denotes the convolution operator, $F$ is the resulting feature map, and $b$ is a bias term.

Several core mechanisms are integrated into the convolutional process:

\begin{itemize}
    \item \textit{Feature Mapping:} Convolution produces feature maps that highlight the activation of specific filters at different spatial locations.
    \item \textit{Weight Sharing:} The same filter is applied across all regions of the input, significantly reducing the number of parameters and enhancing model generalization.
    \item \textit{Nonlinear Activation:} A ReLU activation function is applied to the output of the convolution to introduce nonlinearity.
    \item \textit{Padding and Stride: }Padding preserves spatial dimensions at the borders, while stride controls the filter’s step size during scanning.
\end{itemize}

\subsection{Pre-trained Vision Transformer (ViT)}

The Vision Transformer (ViT) is a transformer-based architecture originally developed for natural language processing and later adapted to vision tasks by representing images as sequences of patches.
\begin{enumerate}
\item \textit{Patch Embedding: }The input image $X$ is divided into $N$ non-overlapping patches, each of which is linearly projected into a $d$-dimensional embedding space:
\begin{equation}
E = \text{Linear}(\text{Patchify}(X))
\end{equation}
where $E \in \mathbb{R}^{N \times d}$ denotes the patch embedding matrix.

\item P\textit{ositional Encoding:} To compensate for the lack of spatial locality in transformers, positional encodings $P$ are added to the patch embeddings to preserve spatial order.

\item \textit{Transformer Encoder:} The sequence of embedded patches is processed by a stack of $L$ transformer encoder blocks. Each block consists of multi-head self-attention and feedforward layers. The output of the $l$-th encoder block is defined as:
\begin{equation}
H^{(l)} = \text{TransformerEncoderBlock}(H^{(l-1)}),
\end{equation}
where \begin{equation} 
H^{(0)} = E + P 
\end{equation}

\item \textit{Classification Head:} The final transformer output $H^{(L)}$ is passed through a linear layer followed by softmax for classification:
\begin{equation}
Y = \text{Softmax}(\text{Linear}(H^{(L)}))
\end{equation}
\end{enumerate}
By pre-training on large-scale datasets like ImageNet, ViT learns rich visual representations. Replacing the classification head with a task-specific layer allows the model to be fine-tuned for fitness movement recognition.

\subsection{Pre-trained EfficientNet}

EfficientNet is a CNN architecture designed to balance accuracy and efficiency through a compound scaling strategy that uniformly scales model depth, width, and input resolution. The pre-trained EfficientNet model, trained on ImageNet, serves as a lightweight feature extractor for various vision tasks.

\begin{enumerate}
\item \textit{Efficient Building Blocks:} The architecture is composed of optimized modules such as depthwise separable convolutions, inverted residual blocks, and squeeze-and-excitation mechanisms, all contributing to reduced computational costs without sacrificing performance.

\item \textit{Transfer Learning:} After pre-training on ImageNet, the classification layers are removed, and the model is fine-tuned with task-specific heads for applications like image classification or movement recognition. This enables EfficientNet to generalize well even on smaller, domain-specific datasets.
\end{enumerate}
\subsection{Pre-trained ResNet50}
ResNet50 is a deep CNN consisting of 50 layers which is based on the Residual Network (ResNet) architecture. It introduces skip connections to mitigate the vanishing gradient problem and enable effective training of very deep networks.

\begin{enumerate}
\item \textit{Residual Blocks:} Each residual block contains convolutional layers with a shortcut connection that bypasses intermediate layers, allowing gradients to propagate more easily and improving convergence.

\item \textit{Bottleneck Design:} ResNet50 employs a bottleneck structure within its residual blocks, consisting of a \(1 \times 1\) convolution, a \(3 \times 3\) convolution, and another \(1 \times 1\) convolution. This design reduces computational cost while preserving model capacity.

\item \textit{Transfer Learning:} Pre-trained on the ImageNet dataset, ResNet50 can be fine-tuned for downstream tasks by replacing its classification head with task-specific layers. This approach enables the reuse of learned visual features on smaller or domain-specific datasets.

ResNet50’s depth and skip connections make it a reliable feature extractor for fitness movement recognition.
\end{enumerate}
\subsection{TimeDistributed Layer}

The TimeDistributed layer applies the same operation independently to each time step in a sequence, making it ideal for tasks like processing video frames. Formally, for an input sequence \( X = \{X_1, X_2, \dots, X_T\} \), the output is computed as:
\begin{equation}
Y_t = f(X_t)
\end{equation}
where \( f \) is the operation applied at each time step \( t \). This layer facilitates efficient parallel computation and is commonly used to connect CNNs to RNNs in video or time-series applications.

\subsection{LSTM Layer}

Long Short-Term Memory (LSTM) is a type of recurrent neural network (RNN) architecture that effectively captures long-term dependencies in sequential data. It is particularly well-suited for tasks that require modeling temporal dynamics, such as video-based human activity recognition. 

An LSTM unit consists of a memory cell that retains information over time and three essential gating mechanisms : the input gate, forget gate, and output gate which regulate the flow of information into, within, and out of the cell. These gates enable selective memory updates and prevent vanishing gradients during training.

In our framework, the LSTM layer processes sequential frame-wise features extracted from video inputs. Its ability to retain relevant historical context across time steps significantly contributes to modeling the temporal structure of fitness movements, leading to improved classification accuracy.

\subsection{Spatial Attention Mechanism}

The temporal attention mechanism enables the network to focus on informative regions of the input while suppressing irrelevant ones.

Given an input feature map $H \in \mathbb{R}^{T \times d}$ from the LSTM layer, attention weights $A \in \mathbb{R}^T$ are computed as:
\begin{equation}
A = \text{Softmax}(W_a \cdot \text{ReLU}(W_h H + b_h) + b_a)
\end{equation}
where $W_h, W_a$ are learnable parameters and $b_h, b_a$ are bias terms.
\begin{itemize}
    \item \textit{Attention Scores:} Assign importance values to each time step in the sequence.  
    \item \textit{Feature Weighting:} Weighted features emphasize relevant spatial cues while downplaying less informative ones.  
    \item \textit{Performance Gain:} By highlighting critical regions, temporal attention improves accuracy and robustness in sequence-based recognition.  
\end{itemize}

\subsection{Dense Layer for Classification}

The Dense (fully connected) layer maps the extracted feature representation to class predictions through a linear transformation followed by an activation function. For an input vector $X \in \mathbb{R}^n$, the output is computed as:
\begin{equation}
Z = W \cdot X + b
\end{equation}
where $W \in \mathbb{R}^{m \times n}$ is the weight matrix and $b \in \mathbb{R}^m$ is the bias.

A softmax activation is applied to $Z$ to generate class probabilities:
\begin{equation}
P_i = \frac{e^{Z_i}}{\sum_{j=1}^C e^{Z_j}}
\end{equation}
where $C$ is the number of classes. The class with the highest probability is selected as the prediction.  

As the final layer of the model, the Dense layer enables supervised learning by mapping high-level features to predefined fitness movement categories.

\section{EXPERIMENTS AND RESULTS}
\subsection{Experimental Setup and Hyperparameters}
All experiments were conducted on a system equipped with an Intel Core i9 11th Gen processor, 32GB RAM, and an NVIDIA RTX A4500 GPU with 36GB of dedicated memory. The proposed model was trained and evaluated on a curated subset of the UCF101 dataset, comprising 35 fitness-related activity classes. For each class, 80\% of the videos were initially allocated to the training and validation set, and the remaining 20\% were reserved for testing. Within the training and validation set, 80\% was used for training and 20\% for validation.

Training was performed using the Adam optimizer with a learning rate of 0.0001, a batch size of 32, and categorical cross-entropy loss. Early stopping was employed to prevent overfitting based on validation loss monitoring.

To compare model efficiency, Table~\ref{param_counts} presents the total, trainable, and non-trainable parameters of each variant.

\begin{table}[h]
\centering
\caption{Parameter Counts of Models}
\label{param_counts}
\footnotesize
\resizebox{\linewidth}{!}{%
\begin{tabular}{|c|c|c|c|}
\hline
\textbf{Model} & \textbf{Total Parameters} & \textbf{Trainable} & \textbf{Non-Trainable} \\ \hline
ViT+LSTM              & \num{87636195}           & \num{215523}          & \num{87420672}        \\ \hline
EfficientNet+LSTM     & \num{9595590}            & \num{9553567}         & \num{42023}           \\ \hline
\textbf{ResNet50+LSTM}& \textbf{\num{27852579}}  & \textbf{\num{27799459}} & \textbf{\num{53120}} \\ \hline
ConvLSTM (Baseline)   & \num{133835}             & \num{133835}          & \num{0}               \\ \hline
\end{tabular}%
}
\end{table}

\subsection{Classification Results}
To evaluate the effectiveness of various spatial-temporal modeling strategies, four deep learning models were implemented. Each model varied in its choice of feature extractor while keeping the temporal modeling and attention components consistent.

\begin{figure}[htbp]
\centering 
\begin{subfigure}[b]{0.48\linewidth} \includegraphics[width=\linewidth]{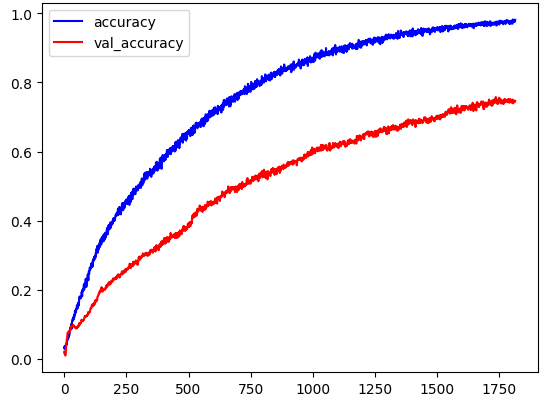} 
\caption{Accuracy vs. Validation Accuracy (LRCN + Attention)} 
\end{subfigure} 
\hfill 
\begin{subfigure}[b]{0.48\linewidth} \includegraphics[width=\linewidth]{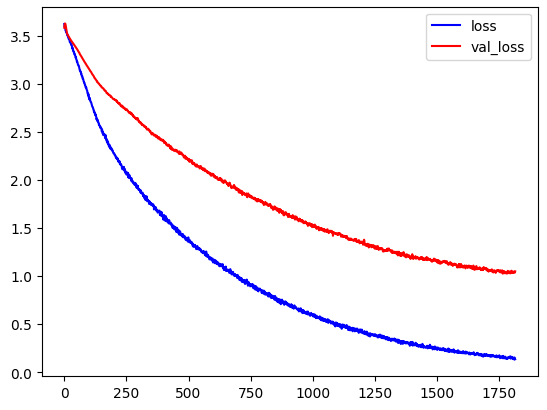} 
\caption{Loss vs. Validation Loss (LRCN + Attention)} 
\end{subfigure} 
\caption{Training Performance Curves of LRCN + Attention Model} 
\label{fig:lrcn_training} 
\end{figure}

\paragraph*{Model 1: LRCN (Conv + LSTM + Attention)}
The baseline LRCN model achieved 73.82\% accuracy and 91.12\% top-5 accuracy. Despite stable training, Fig.~\ref{fig:lrcn_training} shows that its recall and F1 scores were lower due to misclassification of visually similar actions.

\begin{figure}[htbp]
\centering 
\begin{subfigure}[b]{0.48\linewidth} \includegraphics[width=\linewidth]{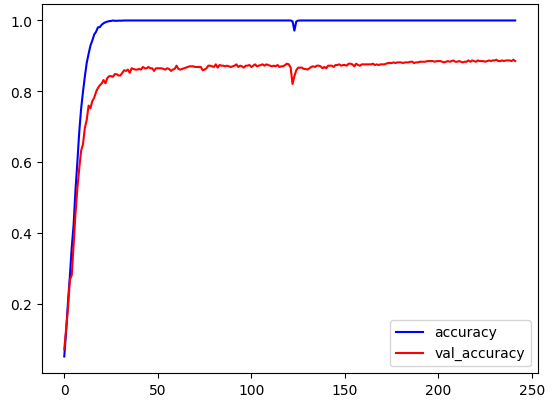} \caption{Accuracy vs. Validation Accuracy (ViT)} \end{subfigure} 
\hfill \begin{subfigure}[b]{0.48\linewidth} \includegraphics[width=\linewidth]{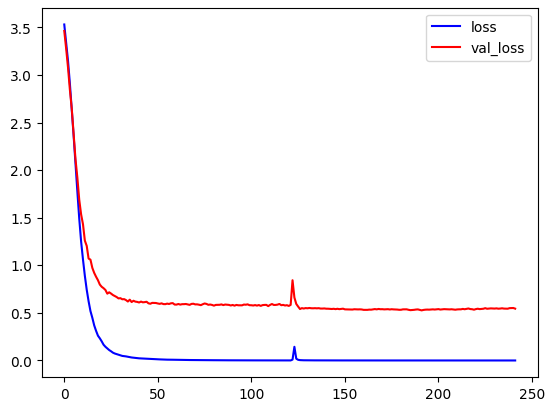} \caption{Loss vs. Validation Loss (ViT)} 
\end{subfigure} 
\caption{Training Performance Curves of ViT + LSTM Model} \label{fig:vit_training} 
\end{figure}

\paragraph*{Model 2: ViT + LSTM}
As shown in Fig.~\ref{fig:vit_training}, incorporating a Vision Transformer significantly improved performance to 88.61\% accuracy. However, the model required higher memory and longer training time due to the transformer complexity.

\begin{figure}[htbp] 
\centering 
\begin{subfigure}[b]{0.48\linewidth} \includegraphics[width=\linewidth]{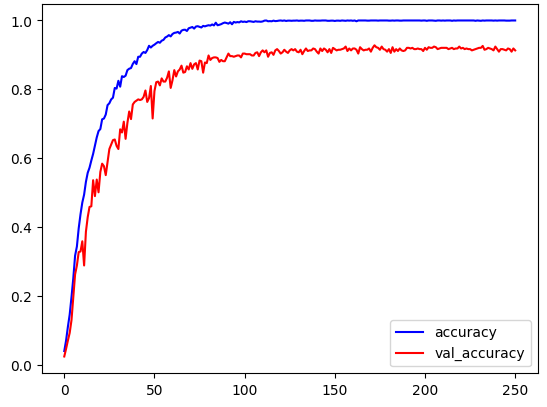} \caption{Accuracy vs. Validation Accuracy (EfficientNet)} \end{subfigure} 
\hfill \begin{subfigure}[b]{0.48\linewidth} \includegraphics[width=\linewidth]{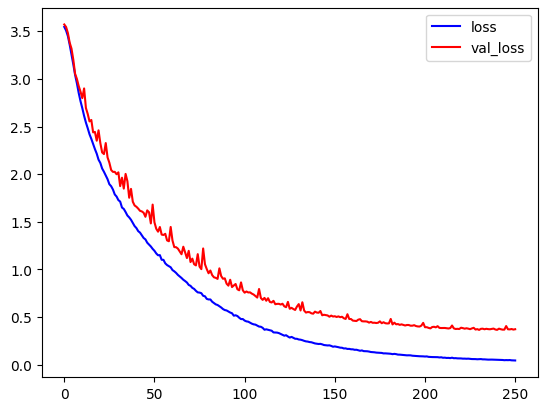} \caption{Loss vs. Validation Loss (EfficientNet)} \end{subfigure} 
\caption{Training Performance Curves of EfficientNet + LSTM + Attention Model} 
\label{fig:eff_training} \end{figure}

\paragraph*{Model 3: EfficientNet + LSTM + Attention}
The EfficientNet-based model achieved an accuracy of 92.01\%. The corresponding training and validation performance trends, including loss and accuracy metrics over epochs, are depicted in Fig.~\ref{fig:eff_training}. 

\begin{figure}[htbp] 
\centering 
\begin{subfigure}[b]{0.48\linewidth} \includegraphics[width=\linewidth]{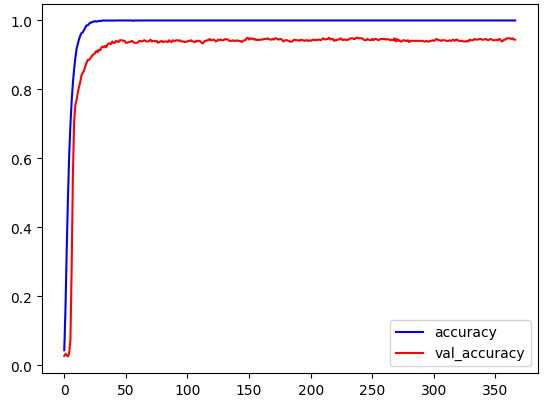} \caption{Accuracy vs. Validation Accuracy (ResNet50)} \end{subfigure} 
\hfill 
\begin{subfigure}[b]{0.48\linewidth} \includegraphics[width=\linewidth]{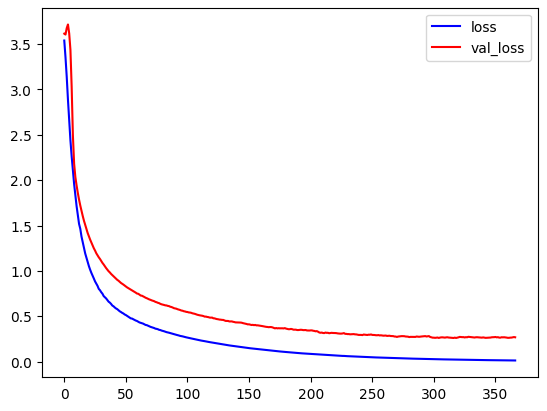} \caption{Loss vs. Validation Loss (ResNet50)} \end{subfigure} 
\caption{Training Performance Curves of ResNet50 + LSTM + Attention Model} 
\label{fig:resnet_training} 
\end{figure}

\paragraph*{Model 4: ResNet50 + LSTM + Attention}
Our best-performing model combined ResNet50 with LSTM and attention, achieving 93.34\% accuracy and 97.04\% top-5 accuracy. As illustrated in Fig.~\ref{fig:resnet_training}, the model maintained stable learning with strong generalization, aided by ResNet’s residual connections and temporal refinement via attention.

Table~\ref{tab:comparative} presents the test set performance comparison across all four proposed models, detailing accuracy, precision, recall, F1 score, and top-5 accuracy.

\begin{table}[htbp]
\caption{Comparative Test Performance of Models}
\label{tab:comparative}
\centering
\footnotesize
\resizebox{\linewidth}{!}{%
\begin{tabular}{|l|c|c|c|c|c|}
\hline
\textbf{Model} & \textbf{Acc.} & \textbf{Top-5 Acc.} & \textbf{Prec.} & \textbf{Recall} & \textbf{F1} \\
\hline
LRCN & 73.82\% & 91.12\% & 75.49\% & 73.82\% & 73.57\% \\
\hline
ViT+LSTM & 88.61\% & 95.71\% & 89.22\% & 88.61\% & 88.53\% \\
\hline
EffNet+Attn & 92.01\% & 97.04\% & 92.37\% & 92.01\% & 91.88\% \\
\hline
\textbf{ResNet50+Attn} & \textbf{93.34\%} & \textbf{97.04\%} & \textbf{93.59\%} & \textbf{93.34\%} & \textbf{93.23\%} \\
\hline
\end{tabular}%
}
\end{table}

Finally, Table~\ref{tab:accuracy_comparison} compares our top-performing model against recent state-of-the-art HAR frameworks from existing literature. It demonstrates that our ResNet50-based model achieves competitive or superior accuracy, validating its effectiveness in recognizing fitness activities.

\begin{table}[htbp]
\caption{Performance Comparison with Existing HAR Approaches}
\label{tab:accuracy_comparison}
\centering
\renewcommand{\arraystretch}{1.2}
\resizebox{\linewidth}{!}{%
\begin{tabular}{|c|c|c|}
\hline
\textbf{Work} & \textbf{Model} & \textbf{Accuracy (\%)} \\
\hline
Chen et al.~\cite{chen2021} & Supervised Hashing (Vision) & 85.60 \\
\hline
Vrskova et al.~\cite{vrskova2022} & Lightweight 3D CNN (Vision) & 89.50 \\
\hline
Zhou et al.~\cite{zhou2022} & ViT + Distillation & 92.90 \\
\hline
Huang et al.~\cite{huang2023} & Dual-GAN (Zero-Shot, Vision) & 89.82 \\
\hline
Bsoul~\cite{bsoul2025} & Graph-Based Skeleton Classifier & 91.80 \\
\hline
Dasan et al.~\cite{dasan2025} & MoViNet-A0 (Mobile Vision) & 91.20 \\
\hline
Maji et al.~\cite{maji2023} & GRU + Attention (Sensor) & 93.20 \\
\hline
\textbf{Ours} & \textbf{ResNet50 + LSTM + Attention} & \textbf{93.34} \\
\hline
\end{tabular}%
}
\end{table}

\section{Conclusion and Future Work}

This paper presents a lightweight framework for fitness movement recognition using video data, combining pre-trained 2D CNNs with LSTM and attention mechanisms to capture both spatial and temporal dynamics. Among the tested configurations, the ResNet50-based architecture with LSTM and attention mechanism achieved the highest accuracy, demonstrating its strength in recognizing fine-grained and repetitive exercise actions. The proposed system contributes a practical solution for real-time activity recognition, with strong applicability in fitness tracking, healthcare, rehabilitation, and coaching. Despite its promising performance, several limitations remain. The model’s accuracy is highly influenced by the diversity and balance of the training dataset, and deep neural networks still pose challenges in terms of computational cost and interpretability.

In our future research, we will focus on expanding the dataset for broader generalization, optimizing the architecture for deployment on resource-constrained devices, and exploring multimodal fusion with wearable sensors. Enhancing user interaction through real-time, explainable feedback is also a key direction to improve accessibility and effectiveness in real-world fitness applications.


\end{document}